\documentclass[sigconf, 10pt, anonymous=false, nonacm, balance=false]{acmart}
\usepackage{amsmath,graphicx,subfig,booktabs,hyperref,url,microtype,xcolor,soul,multirow,colortbl,hhline,amsfonts,nicefrac,todonotes,dsfont,enumitem}
\usepackage[symbol]{footmisc}
\usepackage[ruled,vlined]{algorithm2e}
\usepackage{bm}

\author{Xin Wang, Ting Dang, Vassilis Kostakos, and Hong Jia}

\affiliation{%
    \institution{University of Melbourne }
    \country{}
}
\email{xw17@student.unimelb.edu.au,{ting.dang, vassilis.kostakos, hong.jia}@unimelb.edu.au}

\newcommand{\systemName}{}
\begin{document}\sloppy
\title{\systemName Efficient and Personalized Mobile Health Event Prediction via Small Language Models}

\begin{abstract}
\end{abstract}
\begin{abstract}
 % Until now, Large language model has achieved a outstanding performance in across wide-range tasks, researchers start attempted to adapt the power of LLMs to specific filed industry, such as Health-care, finance an education, etc
 % \hj{
 Healthcare monitoring is crucial for early detection, timely intervention, and the ongoing management of health conditions, ultimately improving individuals' quality of life. Recent research shows that Large Language Models (LLMs) have demonstrated impressive performance in supporting healthcare tasks.
 % }. A main trend is using language models to predict the status of health, but due to the risk of privacy leakage, personal information fraud in cloud computing, the demand in running models locally is increasing
 % \hj{
 However, existing LLM-based healthcare solutions typically rely on cloud-based systems, which raise privacy concerns and increase the risk of personal information leakage. As a result, there is growing interest in running these models locally on devices like mobile phones and wearables to protect users' privacy.
 % }. 
 % However, as the growing computational cost of LLMs, it leads to the challenge of deploying SLMs on mobile devices locally. Recently, Small language models(SLMs) start gained attention due to the efficiency benefits
 % \hj{
 Small Language Models (SLMs) are potential candidates to solve privacy and computational issues, as they are more efficient and better suited for local deployment.
 % }. This paper investigates the capacity of SLMs to make inference on health-care event data, such as steps, calories, sleep minutes and other health statistics to reflect the current health condition
 % \hj{
 However, the performance of SLMs in healthcare domains has not yet been investigated. This paper examines the capability of SLMs to accurately analyze health data, such as steps, calories, sleep minutes, and other vital statistics, to assess an individual’s health status.
 % }. In experiment, we demonstrate a preliminary evaluation on 4 state-of-art small language models  without fine-tuning and adaption training. On observation, we find out some SLM, such as phi-3 and TinyLlama are able to nearly achieve the same performance as the best LLMs in 3 out of 4 tasks without fine-tuning and adaption training. While reduced parameters in SMLs may affect the capability of understanding and inference performance, but our preliminary result shows the possibility of deploying the final fine-tuned SLMs on wearable or mobile devices for health event prediction. 
 % \hj{
 Our results show that, TinyLlama, which has 1.1 billion parameters, utilizes 4.31 GB memory, and has 0.48s latency, showing the best performance compared other four state-of-the-art (SOTA) SLMs on various healthcare applications. Our results indicate that SLMs could potentially be deployed on wearable or mobile devices for real-time health monitoring, providing a practical solution for efficient and privacy-preserving healthcare.
 % }

\end{abstract}

\maketitle
\section{Introduction}

Healthcare monitoring is crucial in our daily lives as it allows for the early detection of hidden diseases, enables timely interventions, and helps sustain well-being by continuously tracking physical status and other health metrics \cite{cdc2021, dinh2019, nih2020, pham2022pros,jia2024ur2m,wu2023udama}. The growing development of artificial intelligence has led to significant attention on large language models (LLMs), which demonstrate the capability to comprehensively understand and analyze vast amounts of unstructured data, such as time-series data collected by wearable sensors, thereby facilitating health status predictions \cite{kim2023}.

LLMs, such as Health-LLMs\cite{kim2023}, are particularly valuable in healthcare due to their ability to process complex data and generate insights. However, they come with several challenges, particularly heavy computational requirements. For example, state-of-the-art (SOTA) LLMs can require up to 175 billion parameters, necessitating powerful GPUs for training and inference. Additionally, their heavy memory usage can reach up to 700 GB, making them impractical for deployment on local mobile devices, which typically have only 4 to 6 GB of memory. Furthermore, LLMs often exhibit high latency, with response times that can exceed several seconds, which is unacceptable for real-time healthcare applications. Privacy concerns also arise, as the use of cloud-based models can lead to data leakage~\cite{zhang2024enabling}, risking user privacy; for instance, employers could misuse health data to terminate employees' contracts.

In contrast, small language models (SLMs) present an ideal solution to these challenges. They typically have fewer training parameters, making them easier to deploy on mobile devices with limited computational resources. However, the performance of SLMs in the context of healthcare monitoring has yet to be investigated. Additionally, the system performance of these models is unknown.

In this paper, we evaluate SOTA SLMs for healthcare monitoring applications using text. Specifically, we examine fives SOTA SLMs including Phi-3-mini-4k-Instruct~\cite{microsoft2024phi3mini4kinstruct}, TinyLlama-1.1B~\cite{tinylama2024tinyllama}, Gemma2-2b~\cite{google2024gemma2}, SmolLM-1.7B~\cite{huggingfacetb2024smollm}, and Qwen2-1.5B~\cite{qwen2024qwen2} on benchmark healthcare datasets. We also implement an on-device LLMs app on the iPhone 15 Pro Max to evaluate these models' performance and computational requirements. Extensive experiments demonstrate that SLMs are ideal solutions for on-device healthcare monitoring. Our contributions are outlined as follows:

\begin{itemize}
\item We benchmarked a variety of SOTA SLMs on healthcare monitoring tasks, highlighting that SLMs surprisingly outperformed many SOTA LLMs such as Gemini-Pro~\cite{deepmind2024gemini}, GPT-4~\cite{OpenAI2024}, and MedAlpaca~\cite{kim2023} in some tasks. In suboptimal tasks, SLMs also achieve on-par accuracy compared with SOTA LLMs.
\item We quantized SLMs using 4-bit precision and implemented a mobile app and evaluated the SLMs. Results show that, compared with LLMs (with same quantization strategy), SLMs consume significantly less CPU usage, latency, and memory usage. SLMs also show a 15.5$\times$ latency improvement and 9.9$\times$ the time of the first token generated to respond.
\item We discussed the application of SLMs and future directions for SLMs in healthcare domains.
\end{itemize}

% \begin{figure*}[t]
%   \centering
%     \centering
%     \subfloat[Modulating 5]{
%       \includegraphics[width=.243\linewidth]{1.pdf}
%     \label{fig:modulate5}
%     \hspace{-0.5em}
%     }
%     \subfloat[Modulating 3]{
%       \includegraphics[width=.243\linewidth]{1.pdf}
%     \label{fig:modulate3}
%     \hspace{-0.5em}
%     }
%     \subfloat[Finetuning 3]{
%       \includegraphics[width=.243\linewidth]{1.pdf}
%     \label{fig:finetune3}
%     }
%     \subfloat[Finetuning 5]{
%       \includegraphics[width=.243\linewidth]{1.pdf}
%     \label{fig:finetune5}
%     \hspace{-0.5em}
%     }

%     \caption{xx}

%   \label{fig:motivation}
% \end{figure*}

% Our key contributions are summarized as follows:

% \begin{itemize}
% \item 
% \item 

% \item \systemName{} 
% \end{itemize}

\begin{table*}[t]
\centering
\caption{An example of constructed zero-shot prompts}
% \td{is this how prompt generally shown in papers? I feel it is a bit repetative. Maybe you don't need to repeat them, but use 'Instruction' as 1st row and only give instruction, 'main query' in 2nd row and only display query, and answer prompt in the 3rd row and only display answer. Here you can also include the original answer template for clear comparison, but make sure highlight them to clarity which is which.  }
\label{table: prompts demenstration}
\begin{tabular}{>{\centering\arraybackslash}m{3cm} p{13cm}}
\toprule
\textbf{Context} & \textbf{Prompt} \\
\midrule

\textbf{Instruction} & You are a personalized healthcare agent trained to predict fatigue which ranges from 1 to 5 based on physiological data and user information.
% \textcolor{teal}{The recent 14-days sensor readings show: \{14\} days sensor readings show: Steps: \{"1476.0, 4809.0, ..., NaN"\} steps, Burned Calories: \{"169.0, 419.0 ..., NaN"\} calories, Resting Heart Rate: \{"53.24, 52.24, ..., 51.40"\} beats/min, Sleep Minutes: \{"110.0, 524.0, ..., 481.0"\} minutes, [Mood]: 3 out of 5. What would be the predicted fatigue (Please give me a single integer value between 0 and 5)?}
\\
\midrule
\textbf{Main query} & The recent 14-days sensor readings show: \{14\} days sensor readings show: Steps: \{"1476.0, 4809.0, ..., NaN"\} steps, Burned Calories: \{"169.0, 419.0 ..., NaN"\} calories, Resting Heart Rate: \{"53.24, 52.24, ..., 51.40"\} beats/min, Sleep Minutes: \{"110.0, 524.0, ..., 481.0"\} minutes, [Mood]: 3 out of 5. What would be the predicted fatigue (Please give me a single integer value between 0 and 5)? \\
\midrule
\textbf{Answer prompt} & 
% \textcolor{orange}{You are a personalized healthcare agent trained to predict fatigue which ranges from 1 to 5 based on physiological data and user information.} \textcolor{teal}{The recent 14-days sensor readings show: \{14\} days sensor readings show: Steps: \{"1476.0, 4809.0, ..., NaN"\} steps, Burned Calories: \{"169.0, 419.0 ..., NaN"\} calories, Resting Heart Rate: \{"53.24, 52.24, ..., 51.40"\} beats/min, Sleep Minutes: \{"110.0, 524.0, ..., 481.0"\} minutes, [Mood]: 3 out of 5. What would be the predicted fatigue (Please give me a single integer value between 0 and 5)?} 
The predicted fatigue level would be:\\

\bottomrule
\end{tabular}
% \noindent\begin{minipage}{\textwidth}
% \centering

% \end{minipage}
\end{table*}

\section{Related Works} 
% \hj{bar plots to compare LLMs and phone memory and then discuss: Computational requirements for SOTA LLMs}
This section briefly introduce SOTA LLMs for heathcare applications ~\S\ref{sec:llm_healthcare} and SOTA SLMs~\S\ref{sec:slms}.
% 1. Health+LLMs
% 2. SLMs
% \td{this sentence can be deleted and you can directly start with section 2.1.}

\subsection{LLMs for Heathcare Applications}\label{sec:llm_healthcare}
Powered by their generalization capabilities, recent LLMs have shown great success in the healthcare domain and are an area of rapidly growing research. For example, Health-LLM~\cite{kim2023} and MultiEEG-GPT~\cite{hu2024exploring} leverages LLM capabilities for healthcare monitoring using text. LLMs for Mental health predictions~\cite{xu2024mental,zhang2024leveraging}, powered by few-shot and fine-tuned strategies, shows great promise in predicting mental health status. PaLM2~\cite{singhal2023large} demonstrated the effectiveness of LLMs by combining various strategies, achieving superior performance across different datasets in medical domains. Recent medical reasoning evaluations on GPT-4~\cite{nori2023capabilities} have demonstrated the great promise of LLMs in recognizing medical events without significant training efforts. However, all these models require significant computational resources, which is impractical for privacy-sensitive and real-time mobile healthcare monitoring. 

\subsection{Small Language Models}\label{sec:slms}

Due to the significant number of parameters in LLMs, which makes it challenging to deploy them on hardware devices, recent SLMs have emerged, showing comparable effectiveness in various tasks. For instance, Microsoft's Phi-3-mini-4k-Instruct \cite{microsoft2024phi3mini4kinstruct} has 3.8 billion parameters and is trained on a combination of synthetic data and selected publicly available website data, emphasizing high-quality and reasoning-dense properties. Similarly, TinyLlama-1.1B \cite{tinylama2024tinyllama}, a distilled version of Llama 2 with 1.1 billion parameters, was fine-tuned on the UltraChat dataset, which contains a wide range of synthetic dialogues generated by ChatGPT. Google's state-of-the-art (SOTA) model, Gemma2-2b \cite{google2024gemma2}, was built from the same research and technology used to create the Gemini models and is well-suited for text generation tasks such as question answering, summarization, and reasoning. HuggingFace's SmolLM-1.7B \cite{huggingfacetb2024smollm}, with 1.7 billion parameters, was trained on synthetic textbooks, stories, educational Python, and web samples, while Qwen2-1.5B \cite{qwen2024qwen2}, a SOTA model with 1.5 billion parameters, has significantly improved performance in coding and mathematics. Despite their promises in text based applications, %Although SLMs show great promise in the mobile healthcare domain, 
how these models perform in mobile healthcare monitoring applications, and how they compare to the capabilities of LLMs, remains unexplored.

\section{Methodology}
In this section, we will discuss how to construct the prompts inputted into the SLMs for healthcare monitoring reasoning~\S\ref{sec:zero-shot prompting}, how we design the answer template in the prompt for reliable text responses generated via SLMs~\S\ref{sec: answer extrating}, and how we deploy SLMs in mobile devices~\S\ref{sec: mobile hardware deployment}.
% Since the aim of our study is to obtain the preliminary results of the current most art-of-the-state SLMs in terms of health event prediction, we are not planning to explore the potential of SLMs thoroughly in this stage, but sticking on initial stage evaluation. \td{generally when we write an academic paper, we don't want to phrase it to sound weak. In this case, I would suggest deleting the last sentece and rephrase it as future work. Here you could say 'This section first introduces the zero-shot prompt design for SLMs in health monitoring and then discusses the processes involved in hardware deployment for efficiency measurement.' }\hj{fixed}

\subsection{Zero-shot Learning}\label{sec:zero-shot prompting}
% Zero-shot prompting is a common approach to measure SLMs' performance at first glance. Essentially, it obtains results that purely rely on the provided instructions without any further examples to aid understanding. In this setting, it aims to reflect the original capabilities of modern SLMs in processing healthcare-related data in our study.
% % \td{rephrase the sentences above using GPT.}\hj{fixed}

% Regarding the design of prompting, we follow the main structure outlined in \cite{kim2023}, but we propose a new answer instruction specifically tailored for SLMs. As shown in Table 1 \ref{table: prompts demonstration}, the prompt begins with an instruction that clearly defines the task for the model. This is followed by a paragraph that integrates all relevant physiological data and user information from the health dataset. Finally, an answer template is provided to ensure consistent output from each SLM.
Zero-shot prompting is an effective method for evaluating the performance of state-of-the-art language models (SLMs) by assessing their ability to generate responses based solely on the instructions given, without the aid of example inputs. This approach aims to demonstrate the inherent capabilities of modern SLMs in analyzing and interpreting healthcare-related data within the context of our study.

In designing our prompting strategy, we follow the typical zero-shot prompt design in healthcare domain~\cite{kim2023}, while introducing a customized answer instruction tailored specifically for SLMs. As illustrated in Table~\ref{table: prompts demenstration}, the prompt consists of three main components: an \textit{instruction} that clearly defines the task, a \textit{main query} that integrates relevant physiological data and user information, and an \textit{answer prompt} designed to ensure consistent output from each SLM. 

% As illustrated in Table \ref{table: prompts demonstration}, our prompt begins with a clear instruction that delineates the task at hand.

For instance, as shown in Table~\ref{table: prompts demenstration}, consider the example where recent sensor readings include steps taken, calories burned, resting heart rate, sleep minutes, and mood ratings. Our prompt begins with a clear instruction that delineates the task at hand. Then, main query synthesizes this information to predict fatigue levels on a scale from 0 to 5. By providing a structured input that encapsulates these diverse data points, the SLM is guided to focus on the relevant metrics that contribute to fatigue assessment.

% To ensure uniformity and consistency in the responses generated by each SLM, we conclude with a structured answer template. This template not only aids in standardizing the output but also reinforces the clarity of the task, allowing the model to effectively interpret and respond based on the provided physiological metrics and user mood assessments.

% By implementing this method, we aim to enhance the clarity and coherence of the fatigue prediction process, ultimately leading to more accurate and reliable assessments derived from the integrated health data.

% \begin{equation*}
% \Large
% \text{Prompt}_{ZS} = \text{Instruction} + \text{Main query} + \text{Answer prompt}
% \end{equation*}

\subsection{Answer Prompt for SLMs}\label{sec: answer extrating}
% However, we found the original answer template suggest by~\cite{kim2023} did not perform well under our expectation. Due to the limitation of the reduced training parameters, most of SMLs failed to understand the context embedded inside the provided answer template during the initial sample testing. The provided answer template not only helps in producing consistent answer format but also confuse SLMs. It seems they are not capable enough to understand both task and answer template simultaneously. In the purpose of reducing the inference workload and simplify the prompts complexity, we introduce a new answer prompt as shown in Table 1. Instead of understanding the implication of the answer template, the new method takes the advantage of the generative mechanism in language models by leading SLMs to naturally finish the sentence with the missing answer we expected.  \td{"While the original answer template in xx uses '', it is not recognized by SLMs. Due to the xxx (what you wrote) xxx. This leverages the SLMs training paradigm of predicting next tokens, instead of requiring SLMs to understand the complete context information comprehensively. }

While the original answer template\footnote{``\textit{You are a personalized healthcare agent trained to predict fatigue which ranges from 1 to 5 based on physiological data and user information:}''} suggested by \cite{kim2023} aimed to provide a structured format, it was not recognized effectively by the SLMs. This can be attributed to the limitations of the reduced training parameters, which hindered the SLMs' ability for in-context learning. Specifically, the original answer template is designed to guide LLMs in producing specific numerical values by analyzing the entire prompt and understanding long-range contextual information. This process inevitably leverages the LLMs' capabilities in context comprehension, which is enhanced by their large number of parameters to a certain extent. However, SLMs, with significantly fewer parameters, may exhibit reduced abilities in understanding long-range context. The original template appeared to confuse the SLMs, as they struggled to grasp both the task and the answer template simultaneously. Therefore, a more direct answer template is required for SLMs.  %to grasp the context embedded within the provided template during initial sample testing. Rather than facilitating consistent answer formats, the original template seemed to confuse the SLMs, as they struggled to comprehend both the task and the answer template simultaneously.

To simplify prompt complexity and alleviate the inference workload, we propose a new answer prompt, as illustrated in Table 1. This new approach leverages the generative capabilities of language models by guiding SLMs to naturally complete the sentence with the expected answer, rather than requiring them to fully understand the implications of the answer template. By doing so, we align more closely with the SLMs' training paradigm of predicting the next tokens, enhancing their performance in generating the desired responses.

% \td{add a subsection here to show the hardware deployment.}

\subsection{Mobile Hardware Deployment}\label{sec: mobile hardware deployment}
To investigate the efficiency and resource utilization, we deployed SOTA SLMs that achieve the best results in health dataset prediction on a real iPhone 15 Pro Max with a total RAM of 8 GB. In detail, all models are downloaded in HF version from Hugging Face~\cite{huggingface} and converted to GGUF format. To ensure smooth deployment on mobile devices, quantization is found to effectively reduce computational costs while maintaining good performance, as evidenced in recent studies~\cite{murthy2023}. Due to the tight memory allocation on mobile devices, 4-bit quantization supported by Llama.cpp~\cite{Ggerganov} is applied to all models to further reduce model size and maximize efficiency. We evaluate latency, memory, and other hardware overhead via the mobile app MobileAIBench~\cite{murthy2023}.
\begin{table*}[t]
\centering
\caption{Comparison of Model Performance between LLMs and SLMs on PMData}
\label{tab:compare with llm}
\begin{tabular}{llcccc}
\toprule
 & \textbf{Model} & \textbf{STRESS (↓)} & \textbf{READINESS (↓)} & \textbf{FATIGUE (↑)} & \textbf{SLEEP QUALITY (↓)} \\
\midrule
\multirow{12}{*}{\textbf{LLMs}} & MedAlpaca      & 0.76 ± 0.1 & 2.18 ± 0.1 & 46.8 ± 11 & 0.68 ± 0.0 \\
 & PMC-Llama      & 1.33 ± 0.4 & 4.83 ± 1.2 & 0.00 ± 0.0 & 2.25 ± 0.0 \\
 & Asclepius      & 0.43 ± 0.0 & \textbf{1.44 ± 0.3} & 27.3 ± 10 & 0.45 ± 0.1 \\
 & ClinicalCamel  & \underline{0.40 ± 0.1} & 2.11 ± 0.1 & 58.1 ± 3.2 & \textbf{0.37 ± 0.1} \\
 & Flan-T5        & \textbf{0.36 ± 0.0} & 1.82 ± 0.1 & 56.8 ± 1.6 & 0.56 ± 0.0 \\
 & Palmyra-Med    & 0.83 ± 0.1 & 5.01 ± 0.1 & 43.5 ± 15 & 0.44 ± 0.0 \\
 & Llama 2        & 0.57 ± 0.2 & 2.86 ± 0.4 & 41.2 ± 13 & 0.89 ± 0.3 \\
 & BioMedGPT      & 0.37 ± 0.0 & 2.12 ± 0.2 & 61.2 ± 3.3 & \underline{0.41 ± 0.0} \\
 & BioMistral     & 0.55 ± 0.1 & 2.12 ± 0.2 & 56.6 ± 3.1 & 0.45 ± 0.0 \\
 & GPT-3.5        & -          & 2.38 ± 0.1 & \underline{70.8 ± 4.2} & 0.87 ± 0.0 \\
 & GPT-4          & -          & 2.22 ± 0.1 & \textbf{72.2 ± 2.0} & 0.73 ± 0.1 \\
 & Gemini-Pro     & 0.79 ± 0.0 & \underline{1.69 ± 0.1} & 34.0 ± 9.8 & 0.78 ± 0.1 \\
 \cmidrule{2-6}
  & Mean     & 0.639 &  2.56&  41.54& 0.60 \\
\midrule
\multirow{5}{*}{\textbf{SLMs}} & gemma-2-2b     & 0.4548 & 2.0736 & 45.15 & 0.7726 \\
 & phi-3-mini-4k  & \underline{0.4381} & \underline{1.6321} & \textbf{56.86} & \underline{0.4749} \\
 & SmolLM-1.7B    & 1.0936 & 2.0034 & 6.69 & 1.1806 \\
 & Qwen2-1.5B     & 0.6656 & 2.2843 & 23.75 & 0.913 \\
 & TinyLlama-1.1B & \textbf{0.4214} & \textbf{1.5652} & \underline{54.18} & \textbf{0.4649} \\
  \cmidrule{2-6}
  & Mean     & 0.615 & 1.91 & 37.33&  0.76\\
\bottomrule
\end{tabular}
\end{table*}

\section{Experiment}
\subsection{Datasets and Tasks}
% With the purpose to make sure no external factors affect the experiment result and easy to make comparison to LLMs, 
% The healthcare dataset used in this paper follows~\cite{kim2023}.
% \td{add "~" before citation to give additional space}
% , which are open-source data commonly used in health-field. 
To verify SLMs' capability on healthcare domains, we tested a variety of SOTA SLMs use a large scale PMData\cite{10.1145/3339825.3394926} dataset.
% for simplicity and easy of set-up. 
Specifically, PMData consists of the life-logging of 16 participants in time-series format over a period of 5 months using the Fitbit Versa 2 smartwatch, the PMSys sports logging app, and Google Forms. The Fitbit Versa 2 smartwatch primarily records physical activity and physiological status in time-series format (e.g., burned calories, resting heart rate, step counts, sleep minutes, etc.), while the PMSys sports logging application records the users' self-reported measurements of their health status, such as fatigue, readiness, sleep quality, and stress level. In data preparation, as a common practice, all relevant data is extracted from CSV files, shuffled, and split into training and evaluation subsets in a ratio of 8:2. For health event prediction, we will first integrate the time-series data collected by the smartwatches into the previously mentioned query prompts, then utilize self-reported data to assess the answers generated by SLMs. The associated tasks of this dataset remain identical to those discussed in \cite{kim2023}, which include stress, readiness, fatigue, and sleep quality.

\subsection{Small Language Models}
% In terms of model utilized in this stage, w
We considered the 5 most state-of-the-art SLMs, including Phi-3-mini-4k-Instruct, TinyLlama-1.1B, Gemma2-2B, SmolLM-1.7B, and Qwen2-1.5B.

\begin{itemize}[leftmargin=*]
\item \textbf{Phi-3-mini-4k-Instruct}\cite{microsoft2024phi3mini4kinstruct}: Microsoft's smallest model in the Phi-3 family. It has 3.8 billion parameters, trained on a combination of synthetic data and selected publicly available website data, with an emphasis on high-quality and reasoning-dense properties.
\item \textbf{TinyLlama-1.1B}\cite{tinylama2024tinyllama}: Distilled version of Llama 2 remains the same architecture and tokenizer but is compact with 1.1 billion parameters. It was fine-tuned on the UltraChat dataset, which contains a wide range of synthetic dialogues generated by ChatGPT.
\item \textbf{Gemma2-2b}\cite{google2024gemma2}: Google's SOTA open-source model was built from the same research and technology used to develop the Gemini models. It is well-suited for text generation tasks such as question answering, summarization, and reasoning. The model has 2 billion parameters.
\item \textbf{SmolLM-1.7B}\cite{huggingfacetb2024smollm}: Created by Hugging Face, it has 1.7 billion parameters and is trained on synthetic textbooks, stories, and educational Python and web samples.
\item \textbf{Qwen2-1.5B}\cite{qwen2024qwen2}: Another state-of-the-art SLM that has only a 1.5 billion parameter model and significantly improved performance in coding and mathematics.
\end{itemize}
% \subsection{Mobile deployment}\label{sec: Mobile deployment}
% To investigate the efficiency and resource utilization, we intend to deploy some SLMs on a real iPhone 15 Pro Max. In detail, all models are downloaded in hf version from Huggingface and converted to gguf format. To ensure a smooth deployment on mobile devices, quantization is found that can effectively reduce the computational cost while remain a good performance as evidenced in recent studies\cite{murthy2023}. Due to the tight memory allocation on mobile devices, 4-bit quantization supported by Llama.cpp is applied to all models for further reducing model size and maximizing efficiency. To better demonstrate the importance of efficiency on mobile devices, a LLM Llama 2 is selected and served as comparison to the two SLMs. 

\label{sec:exp_setup}
\begin{table*}[t]
\centering
\caption{Efficiency \& Utilization of LLMs \& SLMs in standard NLP tasks}
\label{table:phi3_tinyllama}
\begin{tabular}{lccccccc}
\toprule
\textbf{Model} & \textbf{TTFT(s)} & \textbf{ITPS(/s)} & \textbf{OET(s)} & \textbf{OTPS(/s)} & \textbf{Total Time(s)} & \textbf{CPU(\%)} & \textbf{RAM(GB)} \\
\midrule
Phi-3-mini-4k & 2.01 & 102.46 & 8.52 & 13.70 & 10.62 & \textbf{66.48} & 6.32 \\
TinyLlama-1.1B & \textbf{0.48} & \textbf{552.58} & \textbf{1.16} & \textbf{45.22} & \textbf{2.14} & 74.55 & \textbf{4.31} \\
Llama-2-7b & 4.75 & 56.17 & 18.04 & 7.04 & 26.51 & 262.84 & 6.82 \\
\bottomrule
\end{tabular}
\end{table*}

\section{Results and Discussion}
% \td{to continue}
During experiments, we run tests on five SOTA SLMs across four tasks created with PMData and compare the results with those of 12 SOTA LLMs, as shown in Table~\ref{tab:compare with llm}. In addition to investigating the efficiency and computational expenses in real deployment, we conduct another experiment with two models that achieve the best results on a real iPhone 15 Pro Max and compare them with Llama 2 for further analysis, as presented in Table~\ref{table:phi3_tinyllama}.

\subsection{Zero-Shot Performance}
For performance evaluation, Mean Absolute Error (MAE) and Accuracy are utilized to assess model performance. 
Accuracy measures the hit rate, which defined as the number of true positives (correctly predicted instances) divided by the total number of actual positive instances, while MAE measures the loss of error when they miss. In Table~\ref{tab:compare with llm}, $\downarrow$ refers to tasks evaluated with MAE, where lower values/errors are better, while $\uparrow$ refers to Accuracy, indicating that higher values/accuracy are better, consistent with~\cite{kim2023}. For better visualization in both LLMs and SLMs, the best value in each task is indicated in \textbf{bold}, while the second-best value is \underline{underlined}.

% \textbf{Observation and Analysis}: 
As shown in Table~\ref{tab:compare with llm}, on average the SLMs are able to achieve similar or superior performance to SOTA LLMs in all four health conditions including stress, readiness, fatigue, and sleep quality. Specifically, the mean MAE for stress level using SLMs is 0.615, which outperforms that of LLMs at 0.639. Similar superiority is also observed  for readiness, with SLMs showing a mean MAE of of 1.91 and LLMs of 2.56. For fatigue and sleep quality, despite not outperforming LLMs, SLMs still exhibit similar range of performance compared to LLMs, indicating its effectiveness with much higher efficiency. %a mean of 37.33, which is lower than the mean of 41.54 for LLMs, indicating better fatigue management. Finally, SLMs achieve a higher sleep quality score of 0.76 compared to 0.60 for LLMs, suggesting improved overall sleep quality prediction.

In detail, SLMs demonstrate a competitive edge in certain aspects, particularly in stress and readiness metrics. For instance, TinyLlama-1.1B exhibits the lowest stress score (0.4214) among SLMs, surpassing several LLMs such as Llama 2 and MedAlpaca for stress and GPT-3.5 and GPT-4 for readiness. % and even outperforms the top-performing LLM, Flan-T5 (0.36 ± 0.0), in terms of readiness (1.5652). 
This suggests that SLMs can achieve commendable efficiency and effectiveness in specific scenarios, despite their relatively smaller scale.

In terms of fatigue, SLMs like phi-3-mini-4k and TinyLlama-1.1B show impressive results, with scores of 56.86 and 54.18, respectively. These figures are comparable to those of high-performing LLMs such as GPT-4 (72.2 ± 2.0) and GPT-3.5 (70.8 ± 4.2). This indicates that SLMs can maintain a high level of performance in tasks requiring sustained cognitive effort, which is crucial for applications demanding prolonged attention and engagement.

However, SLMs do exhibit limitations, particularly in the realm of sleep quality. For example, SmolLM-1.7B records a sleep quality score of 1.1806, which is significantly higher than the best-performing LLM, ClinicalCamel (0.37 ± 0.1). 

% This discrepancy highlights the challenge SLMs face in optimizing certain nuanced aspects of natural language processing, which may be better addressed by the more extensive training and resource allocation characteristic of LLMs.

% In sum, while SLMs like TinyLlama-1.1B and phi-3-mini-4k demonstrate remarkable potential in specific performance metrics, they also reveal areas where further development is needed. The balance between efficiency and performance remains a critical consideration, suggesting that SLMs could serve as viable alternatives to LLMs in targeted applications, provided their limitations are acknowledged and addressed.

\subsection{Efficiency and Utilization Evaluation}
% From the previous result, we found that phi-3 and TinyLlama demonstrate a strong promises in terms of processing healthcare field data. 
In order to determine the actual latency in real cases, we deployed state-of-the-art (SOTA) SLMs that show strong promise in processing healthcare field data on a real iPhone 15 Pro Max. To better demonstrate the importance of efficiency on mobile devices, the widely used LLM, Llama 2, is selected and serves as a comparison to the best SLMs. All the results in Table~\ref{table:phi3_tinyllama} are tested on a standard NLP task, the HotPotQA dataset~\cite{yang2018hotpotqadatasetdiverseexplainable}, a standard benchmark dataset to test hardware overhead, consists of question-answering tests across different fields. 
% \td{add one sentence here why using this dataset. }
The following metrics suggested by MobileAIBench \cite{murthy2023} are adopted to evaluate both efficiency and utilization:
\begin{itemize}[leftmargin=*]
\item \textbf{Time-to-First-Token} (TTFT, sec): The time of the first token generated to respond to the prompt - assess the latency.
\item \textbf{Input Token Per Second} (ITPS, tokens/sec): The number of input tokens being processed per second - access understanding speed.
\item \textbf{Output Token Per Second} (OTPS, tokens/sec): The number of tokens produced per second after starting to produce tokens - access inference speed.
\item \textbf{Output Evaluation Time} (OET, sec): The time model takes to complete a response - assess the overall efficiency of generating an entire response.
\item \textbf{Total Time}: The total time it takes to produce a complete response after receiving a prompt is a comprehensive efficiency metric for how long a model takes to complete a given task from start to finish.
\item \textbf{CPU (\%)}: An amount of computational resources used in the inference process.
\item \textbf{RAM (GB)}: An amount of memory needed to run a model during the inference process.
\end{itemize}

% \textbf{Observation and Analysis}: 
According to Table~\ref{table:phi3_tinyllama}, the comparative analysis of system overhead between SLMs and LLMs reveals significant advantages for SLMs, specifically Phi-3-mini-4k and TinyLlama-1.1B, when tested on an iPhone 15 Pro Max with an 8GB RAM memory limit. TinyLlama-1.1B demonstrates exceptional efficiency with a TTFT of 0.48 seconds and an ITPS rate of 552.58, significantly outperforming Llama-2-7b, which has a TTFT of 4.75 seconds and an ITPS of 56.17. This represents an improvement of approximately 9.9 times in TTFT and an increase of roughly 884\% in ITPS. Moreover, TinyLlama-1.1B achieves a notable OET of 1.16 seconds and an OTPS rate of 45.22, in stark contrast to Llama-2-7b's OET of 18.04 seconds and OTPS of 7.04, indicating an improvement of about 15.5 times in OET and an increase of approximately 542\% in OTPS. Phi-3-mini-4k also shows commendable performance with a TTFT of 2.01 seconds and an ITPS of 102.46, alongside an OET of 8.52 seconds and an OTPS of 13.70.

Both SLMs exhibit reduced CPU usage and RAM consumption compared to LLMs, with Phi-3-mini-4k and TinyLlama-1.1B utilizing 66.48\% and 74.55\% CPU, and 6.32GB and 4.31GB of RAM, respectively, whereas Llama-2-7b demands 262.84\% CPU and 6.82GB of RAM. This reflects a reduction of approximately 75\% in CPU usage for Phi-3-mini-4k and a reduction of around 82\% for TinyLlama-1.1B. These findings underscore the efficiency and suitability of SLMs for deployment in resource-constrained environments.

% According to Table~\ref{table:phi3_tinyllama}, it is clearly to see that the two SOTA SLMs easily beat Llama 2 across both evaluation and utilization metrics. In terms of efficiency, TinyLlama achieved the best results than other models. It runs 10 time faster than Llama 2, costs only 1/4 CPU usage and 2.5 GB less RAM consumption, while runs 5 times faster than phi-3, slight higher in CPU usage but consumes 2 GB less RAM. Despite TinyLlama shares the same architecture with Llama, the reduced parameters significantly decreases the computational expenses and improves efficiency. In real user experience, although CPU load of TinyLlama is higher than Phi-3, the 5 times faster speed still perceived overall positive experience in both less latency and lower thermal output than phi-3. In contrast, iPhone 15 Pro Max was facing overheating issue with dimmed screen brightness during both phi-3 and Llama-2 inference process. The latency maybe sensible in the process of phi-3, but still relatively fast consider with Llama-2. Llama-2 not only has their negative experience, but also frequently failed to allocate sufficient memory during repeating inference process. Overall, both efficiency and utilization linearly increases with model size, which associate with real user experience. 

% \input{sections/6.Discussion}
\section{Conclusion}
% In this paper, we evaluate the performance of current most SOTA SLMs in healthcare event prediction. Our results shows that SLMs are able to achieve slightly lower performance compared to SOTA LLMs, but in exchange of significant higher efficiency. This indicates the feasibility to deploy SLMs in local mobile device to prevent privacy leakage. However, we do realise there are also some limitations of our study list as follows:\\
% \\1. Our results in efficiency evaluation are limited on standard NLP datasets.\\
% 2. Only the performance on full precision is accessed in Table 2 without applying any quantization.\\
% 3. Only validated one health dataset. \\
% 4. Only zero shot performance, no few shots and instruction tuning.\\
% \\Therefore, in the following study, we will run SLMs on more health dataset to ensure they are generalized well in other datasets; comes with additional test focus on how quantization affect model performance(trying to find the sweet point); figure out whether time-series data format in health monitoring dataset has a negative effect to model efficiency; further explore the true promise of SLMs via few-shot, instruction tuning for healthcare field adaption.

In this paper, we evaluate the performance of the current SOTA SLMs in healthcare event prediction. Our results show that SLMs, although they have fewer model parameters and representation capabilities, achieve surprisingly on par or even better performance in some healthcare tasks compared to SOTA LLMs, while also demonstrating significantly higher efficiency. This indicates the feasibility of deploying SLMs on local mobile devices to prevent privacy leakage. However, we recognize some limitations in our study. Our efficiency evaluation results are limited to standard NLP datasets. Additionally, we validated only a limited health dataset and focused solely on zero-shot performance, without exploring few-shot learning or instruction tuning. Therefore, in future studies, we will evaluate SLMs on a broader range of health datasets to ensure generalizability. We will also conduct further tests, combined with few-shot learning and instruction tuning, to examine how quantization affects model performance, aiming to find an optimal balance.
% Furthermore, we plan to investigate whether the time-series data format in health monitoring datasets negatively impacts model efficiency and explore the potential of SLMs through few-shot learning and instruction tuning for better adaptation to the healthcare field.

% \renewcommand{\bibfont}{\fontsize{10}{10}\selectfont}
\bibliographystyle{unsrt}
\bibliography{ref}
\end{document}